\documentclass[11pt]{article}

\usepackage[final]{acl}
\usepackage{xcolor}
\usepackage{times}
\usepackage{latexsym}
\usepackage{stfloats}
\usepackage{enumitem}
\usepackage{fvextra}
\usepackage{longtable}

\usepackage[T1]{fontenc}

\usepackage[utf8]{inputenc}

\usepackage{microtype}

\usepackage{inconsolata}

\usepackage{graphicx}

\author{Joonatan Laato\textsuperscript{1},
  Veera Schroderus\textsuperscript{2}, 
  Jenna Kanerva\textsuperscript{1}, \\ 
  {\bf Jenni Kauppi\textsuperscript{2},
  Virpi Lummaa\textsuperscript{2}, \and
  Filip Ginter\textsuperscript{1,3}} \\
  \textsuperscript{1}TurkuNLP, University of Turku, Finland \\
  \textsuperscript{2}Department of Biology, University of Turku, Finland\\
  \textsuperscript{3}ELLIS Institute Finland \\
  \texttt{\{joonatan.m.laato, veera.i.schroderus,}\\\texttt{jmnybl, jejakau, virpi.lummaa, figint\}@utu.fi} \\
  }
  
\usepackage{float}  
\title{Measuring Social Integration Through Participation: Categorizing Organizations and Leisure Activities in the Displaced Karelians Interview Archive using LLMs}

\date{Feb 2026}

\begin{document}

\maketitle

\begin{abstract}

Digitized historical archives make it possible to study everyday social life on a large scale, but the information extracted directly from text often does not directly allow one to answer the research questions posed by historians or sociologists in a quantitative manner. We address this problem in a large collection of Finnish World War II Karelian evacuee family interviews. Prior work extracted more than 350K mentions of leisure time activities and organizational memberships from these interviews, yielding 71K unique activity and organization names---far too many to analyze directly.

We develop a categorization framework that captures key aspects of participation (the kind of activity/organization, how social it typically is, how regularly it happens, and how physically demanding it is). We annotate a gold-standard set to allow for a reliable evaluation, and then test whether large language models can apply the same schema at scale. Using a simple voting approach across multiple model runs, we find that an open-weight LLM can closely match expert judgments. Finally, we apply the method to label the 350K entities, producing a structured resource for downstream studies of social integration and related outcomes.
\end{abstract}

\section{Introduction}

Digitizing large collections of historical records, combined with the use of large language models (LLMs), has opened new possibilities for analyzing cultural and social patterns at a large scale. LLMs make it straightforward to automate tasks such as extracting entities and other structured information from massive archival collections. In many cases, however, entity extraction alone is not sufficient; additional steps---such as grounding and grouping entities---are often needed to support the underlying research.

In this study, we examine a unique historical corpus: \textit{Siirtokarjalaisten tie} (\textit{The Path of Displaced Karelians}), a large interview collection documenting World War II evacuees who were permanently relocated from the Karelia region to different parts of Finland. The archive contains testimonies from approximately 160,000 individuals, offering rare coverage of everyday social life at scale. The interviews are brief (see Figure~\ref{fig:single_story}), but they typically include information about organization membership and leisure activities. Organizations include a wide variety of different clubs, choirs, work unions, boards, and other similar organizations, while leisure activities primarily refer to hobbies, sports, outdoors activities, and similar. Together, these in many cases allow inferring the degree of social integration of the individuals, and, if extracted across the whole data, of the displaced population at large. Extracting this information is the primary focus of this study.

In prior work, we used LLMs to extract the organization membership and leisure activity mentions from the interviews \citep{laato2025extractingsocialconnectionsfinnish}. The resulting dataset contains 354,302 extracted entity mentions from 89,339 interviews. However, without further grounding and interpretation, these entities are not directly usable as a proxy for social integration. For instance, the popular ``Martha'' club of homemakers focusing on spreading household advice appears under multiple free-form variants (e.g., ``Marthas'' vs.\ ``Martha Club''), and historically meaningful groups may be described with widely differing names across regions and dialects. Critically, mere mentions of organizations and leisure activities lack metadata that would help estimate the degree of social integration implied by a given activity.

This motivates a central step between extraction and domain analysis: normalization and categorization. For sociological questions---especially those focused on social integration and its links to health and longevity---researchers need a representation of participation that is comparable across individuals and communities. Achieving this requires mapping historical mentions to a consistent, interpretable classification framework that reflects culturally specific forms of Finnish social life.

To support this downstream analysis, we evaluate whether LLMs can assist with this structuring step and how closely they can match human expert judgment. In collaboration with sociologists (domain experts), we develop a classification framework designed to capture the dimensions of social participation most relevant to studying integration patterns, and we assess both human reliability and model performance within that framework.

We address three primary questions: \begin{enumerate}[noitemsep,topsep=3pt]
    \item Is it feasible to represent the multi-dimensional nature of historical Finnish social organizations and leisure activities in a way that still yields reliable human consensus?
    \item If so, to what extent can LLMs replicate domain-expert categorizations of culturally specific historical entities?
    \item Can iterative and automated prompt engineering further improve model performance on this task?
\end{enumerate}

\section{Data}

Our primary source ``Siirtokarjalaisten tie'' (The Path of Displaced Karelians) is a 4-volume book series documenting Finnish citizens permanently displaced from Eastern Karelia following the 1939-1944 war with the USSR. The volumes contain 89,339 family interviews conducted in 1968--1970 by approximately 300 trained interviewers, representing approximately 160,000 adults from the roughly 420,000 total refugees. All interviews were originally conducted in Finnish. The books were digitized through OCR by \citet{loehr2017newly}.


\begin{figure}[bt]
    \centering
    \includegraphics[width=0.9\columnwidth]{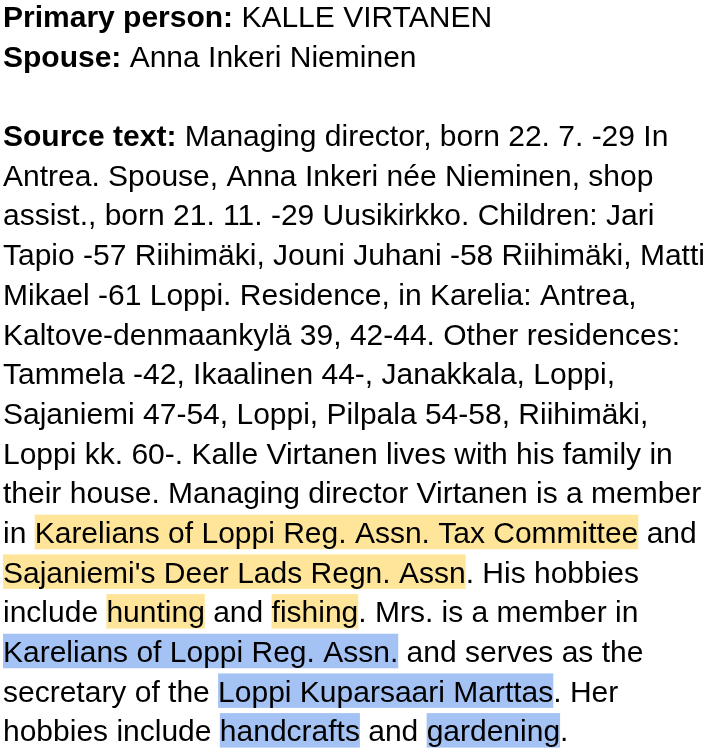}
    \caption{A single interview example with entities pertaining to the husband (yellow) and wife (blue) highlighted. Translated from Finnish, preserving typographic features of the original data. Note: Names and dates have been changed.}
    \label{fig:single_story}
\end{figure}

The interviews combine standardized demographic sections with free-text information about daily life, including social organization membership and hobbies. Figure~\ref{fig:single_story} illustrates an example interview showing the information typically captured. Previously, we used OpenAI's GPT-4-turbo LLM to extract all mentions of social entities from these texts \citep{laato2025extractingsocialconnectionsfinnish}, resulting in 192,388 mentions of hobbies and 161,914 mentions of organizations, extracted with 88.8\% F-score, close to human performance at the task. These mentions comprise a total of 71,874 unique entities (7,674 hobbies and 64,200 organizations), as common activities were mentioned by many individuals. 




\begin{table}[t!]

\centering
\small
\renewcommand{\arraystretch}{1.05}
\setlength{\tabcolsep}{6pt}

\begin{tabular}{|p{\columnwidth}|}
\hline
\textbf{Top 10 Hobbies}\\
\hline
handicrafts (\textit{käsityöt}) — 31{,}780\\
fishing (\textit{kalastus}) — 20{,}222\\
literature (\textit{kirjallisuus}) — 13{,}128\\
gardening (\textit{puutarhanhoito}) — 10{,}519\\
reading (\textit{lukeminen}) — 9{,}324\\
hunting (\textit{metsästys}) — 6{,}287\\
skiing (\textit{hiihto}) — 5{,}912\\
outdoor activities (\textit{ulkoilu}) — 5{,}597\\
sports (\textit{urheilu}) — 4{,}997\\
handicrafts [partitive case] (\textit{käsitöitä}) — 4{,}708\\
\hline
\textbf{Top 10 Organizations}\\
\hline
Karelian Society (\textit{Karjalaseura}) — 6{,}610\\
Martha Association (\textit{Marttayhdistys}) — 5{,}725\\
youth association (\textit{nuorisoseura}) — 2{,}545\\
Farmers' Association (\textit{Maamiesseura}) — 2{,}476\\
Marthas (\textit{Martat}) — 1{,}967\\
Agricultural Women (\textit{Maatalousnaiset}) — 1{,}957\\
Lotta Svärd (\textit{Lotta Svärd}) — 1{,}930\\
sports club (\textit{urheiluseura}) — 1{,}252\\
church choir (\textit{kirkkokuoro}) — 1{,}230\\
elementary school (\textit{kansakoulu}) — 1{,}212\\
\hline
\end{tabular}

\caption{Top 10 most frequently mentioned hobbies and organizations. Finnish terms in italics.}
\label{tab:top10_both}
\end{table}

\begin{figure}[h]
    \centering
    \includegraphics[width=\columnwidth]{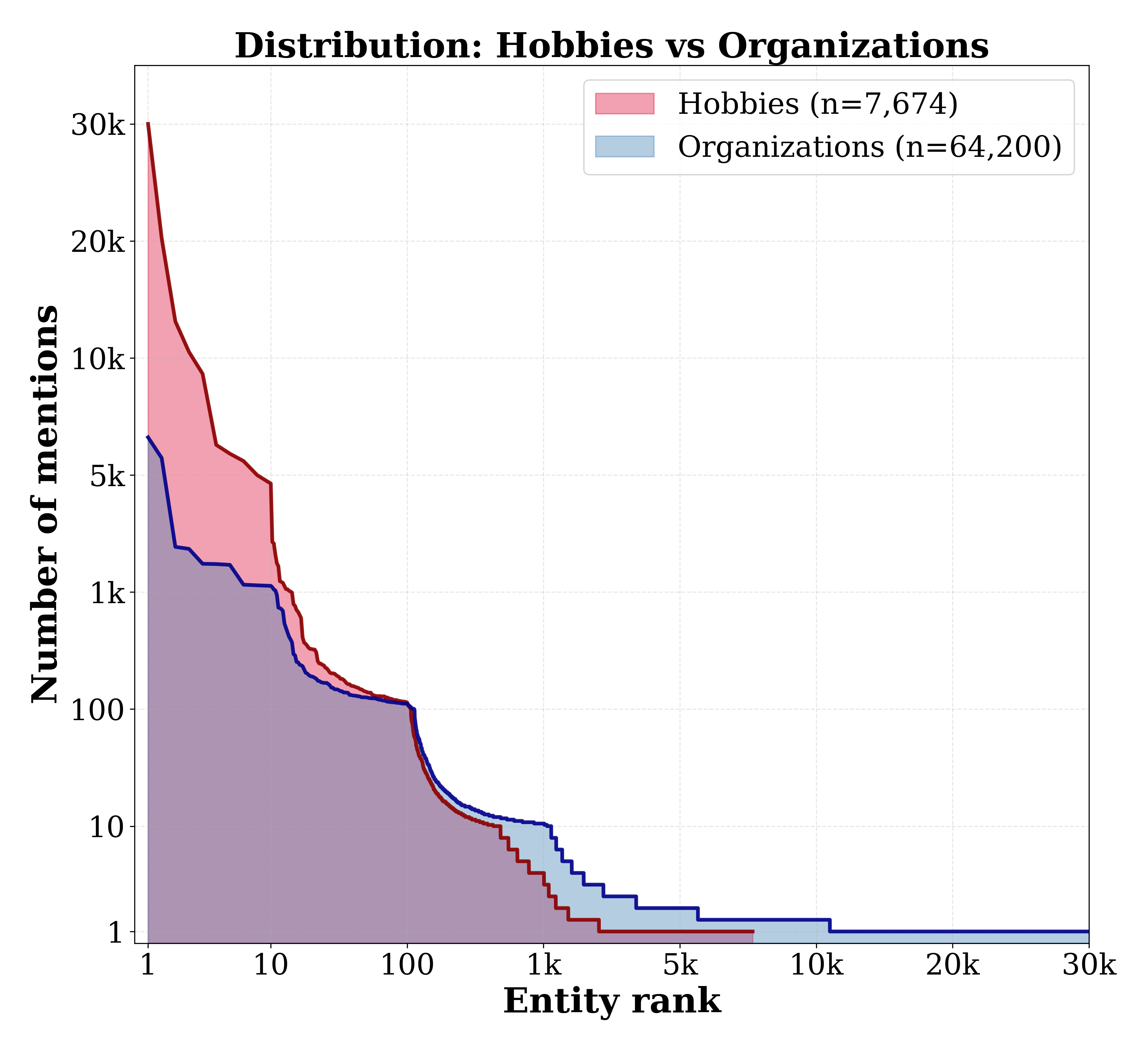}
    \caption{Distribution of hobby and organization mentions across entity ranks. Hobbies (red) show concentrated distribution in the first few mentions, with some 7.6k unique entities, while organizations (blue) show a long tail extending beyond 60k unique entities.}
    \label{fig:distribution}
\end{figure}

The top 10 most frequently mentioned entities (Table~\ref{tab:top10_both}) already show very active lifestyle and organizational involvement of the Karelian population in post-war Finland. Common hobbies center on traditional rural activities such as handicrafts, fishing, and gardening. Meanwhile organizations include Karelian cultural societies, women's associations, and agricultural groups. 

As illustrated in Figure~\ref{fig:distribution}, the extracted hobbies and organizations display unsurprisingly different distribution. Hobbies include only 7,674 unique entities, where the top 200 account for 92.6\% of all mentions (192,388 total). In contrast, organizations display a long-tail distribution with 64,200 unique entities, where the top 200 capture only 49.0\% of all mentions (161,914 total). 

\subsection{Entity Normalization Challenges}

Pattern matching reveals variations referring to effectively the same activity. The term ``mart*'' matches 156 unique organization strings (out of 64,200 total unique organizations) including: Martha Association (\textit{Marttayhdistys}), Marthas (\textit{Martat}), marthas (\textit{martat}), Martha Society (\textit{Marttaseura}), Martha Club (\textit{Marttakerho}). The same root appears in 40 unique hobby strings (out of 7,674 total unique hobbies) including: Martha work (\textit{marttatyö}), Martha activities (\textit{marttatoiminta}). These all for the most part refer to the same popular homemaker organization and its activities.

The pattern ``karjala*'' (Karelia) yields 590 unique organizational strings referring to various societies established by the Karelians, often to promote their interests and maintain connections, and 94 unique hobby strings predominantly describing Karelia-specific food preparation: e.g.\ making Karelian pastries (\textit{karjalanpiirakoiden valmistus}).

Handicraft variations (``käsit*'') show 78 unique hobby strings: handicrafts (\textit{käsityöt}), handicrafts-partitive (\textit{käsitöitä}), making handicrafts (\textit{käsitöiden tekeminen}), handicraft making (\textit{käsityöiden tekeminen}). The pattern also matches 22 unique organizational strings: handicraft club (\textit{käsityökerho}).

Additionally, many organizations with entirely different names belong to the same functional category. For example, various sports clubs while having unique names should nevertheless be grouped together for the analysis. 

\subsection{Pre-processing: Hierarchy Extraction}

We pre-process the organizations using an LLM to separate out hierarchies of organizations and individual roles: given an entity like ``Chairman of the Administrative Board of the Karelian Society,'' we prompt an LLM (Llama-3.3) to separate the individual's role as well as the underlying hierarchy. In this case we would extract ``Chairman'' as a role and ``Administrative Board'' as a hierarchical element. The remaining entity is then interpreted as the actual social organization, here ``Karelian Society''. This reduced the unique social organization count from 64,200 to 52,982, extracting 2,147 different roles and 709 unique hierarchies.


\section{Related Work}

Historical and cultural data presents unique challenges for entity normalization and clustering: (1) classifying entities into meaningful groups, (2) normalizing variant names, and (3) capturing semantic relationships through multi-dimensional categorization.

Previously, these have been approached using distributional representations such as word2vec and fastText \citep{mikolov2013efficient,bojanowski2017fasttext} to capture semantic similarities through vector clustering \citep{baroni2014don}. Transformer-based models like BERT \citep{devlin2019bert} advanced multi-label classification through contextualized representations, but require substantial domain-specific labeled data for training.
More recently, LLMs have enabled clustering entities based on contextual embeddings. \citet{huang2024textclustering} transform text clustering into classification by having the LLM generate potential labels, merge semantically similar labels, and assign entities accordingly. This label-merging approach captures underlying data structure without requiring any fine-tuning. \citet{viswanathan2024large} demonstrated few-shot clustering through LLM-generated pairwise constraints and entity canonicalization with minimal examples. Recent work on entity matching \citep{entitymatching2024} highlights how LLMs implement semantic equivalence beyond syntactic similarity, though challenges remain for culturally-specific entities and abbreviated names.
However, multi-label classification with LLMs presents distinct challenges. \citet{ma2025llms} found that LLMs suppress multiple labels during generation despite applicable categories. \citet{lan2024multi} addressed this through multi-label contrastive learning. \citet{tabatabaei2024llm} demonstrated successful industrial-scale deployment with custom taxonomies for hierarchical multi-label classification.
Across these tasks, careful prompt engineering significantly improves LLM's classification accuracy, with automated methods like APE \cite{zhou2022large} and OPRO \cite{yang2024largelanguagemodelsoptimizers} emerging to generate and iteratively optimize prompts.

\section{Methods}

While recent work shows that LLMs can cluster semantically \cite{huang2024textclustering} and normalize entities \cite{viswanathan2024large}, these approaches focus on single-label classification or binary similarity judgments. However, as demonstrated in our data analysis, the diversity of historical social organizations resists discrete categorization. Attempting to single-label these entities would require hundreds of narrowly-defined categories. Instead, a multi-label approach that captures specific dimensions of each entity proves more appropriate. Furthermore, rather than forcing data into pre-existing taxonomies (as is commonly used in NER), our approach develops classification dimensions directly towards sociological research questions about social integration patterns.


\subsection{A Classification Framework for Social Integration Research}

The design of this framework is directly guided by the main research objective of sociologists: characterizing the social integration of Karelian refugees after relocation. Combined with further connected data on this population, such as health and mortality information, this characterization of social integration can be used to answer questions about how different levels of integration influenced mortality outcomes. Furthermore, interactions of social integration level can be measured relative to numerous factors, such as age, residential history, distance from relatives, and whether the relocation occurred individually or with the community. The dataset's scale and demographic uniformity provide a rare opportunity to isolate the effects of post-migration social environments on individual outcomes.

While this data, and approach, will eventually be utilized broadly across different fields, initially, the main focus will be on investigating the health effects of different kinds of social integration. Consequently, key questions and category types of the annotation framework were constructed based on well-known health factors, such as physical activity \cite{iwasaki2001, rhodes2017}, sociality \cite{holtlunstad2010, hansdottir2022}, and mental stimulation \cite{yates2016, hansdottir2022, osullivanmcquade2023}. The frequency of activity is relevant to health in both physical activity \cite{iwasaki2001, rhodes2017} and mentally stimulating leisure activities \cite{weziakbialowolska2023}. For non-solitary activity, frequency and group size additionally reflect the social intensity of participation. 




These considerations led to the design of an annotation schema with four main aspects for each activity/organization, formulated into four questions: \textbf{Q1 - Categories}: The general type(s) of organization/activity; \textbf{Q2 – Group Size}: How many people typically participate? \textbf{Q3 – Regularity}: How often do they meet/engage? \textbf{Q4 – Physical Intensity}: What level of physical activity is involved? Each question has predefined answer options, detailed in Table~\ref{tab:questionnaire}, with special options for ambiguous cases: ``Cannot be determined'' applied when the judgment is impossible to make based on the information given, and ``Data error'' applied when the entity is not, in fact, a leisure-time activity or organization, but rather an extraction error.

To support culturally informed annotation, a questionnaire was distributed through the Karjalan Liitto ry Facebook page, where Karelian migrants and their descendants provided ratings and descriptions for approximately 60 common ambiguous entities. While these responses were not directly incorporated into the current annotation framework, they represent a valuable reference point in the refinement of the guidelines.

\begin{table}[t]
\centering
\small
\begin{tabular}{|p{0.40\columnwidth}|p{0.50\columnwidth}|}
\hline
\textbf{Q1 -- Categories} & \textbf{Q2 -- Group Size} \\

Professional/Work   & \quad Alone \\
Cultural/Traditional         & \quad Small group \\
Religious/Spiritual          & \quad Large group \\
Sports/Physical activity     & \quad Non-definable \\
Creative/Artistic            & \quad Data error \\
Educational/Academic         & \textbf{Q3 -- Regularity} \\   

Social welfare               & \quad Regular \\
Administrative               & \quad Occasional \\
Political                    & \quad Continuous \\
General social group         & \quad Event-based \\
Nature-related               & \quad Non-definable \\
Health-related               & \quad Data error \\

Property/ownership           & \textbf{Q4 -- Physical Intensity} \\  

Animal-husbandry       & \quad Intense \\
Cooking                      & \quad Continuous \\
Military-related             & \quad Light \\
Special Hobbies               & \quad Stationary \\
Non-definable                & \quad Non-definable \\
Data error                   & \quad Data error   \\
\hline
\end{tabular}
\caption{Questionnaire schema showing all options for each question in our final round of annotations.}
\label{tab:questionnaire}

\end{table}

\subsection{Annotation Process}

Four annotators, two domain experts and two machine learning specialists, annotated five iterative rounds of 50 entities, split evenly between hobbies and organizations. After each round the schema and guidelines were refined. For example, annotators should distinguish between organizations and their administrative bodies, separate ``Administrative'' (governance) from ``Political'' (party activities), and select ``Cannot be determined'' rather than guess when uncertain. Once a sufficient agreement had been reached, a set of 200 entities was annotated by all four annotators, and split into a development/evaluation set of 50 entities, and a final test set of 150 entities.

We measured inter-annotator agreement using pairwise Cohen's Kappa across both evaluation (n=50) and test (n=150) sets (See Appendix Table~\ref{tab:iaa_combined} for full results). The pairwise Kappa score shows moderate-to-substantial average agreement of 0.704 (evaluation set) and 0.690 (test set). Annotator agreement is also a proxy for the difficulty of making the judgements, and the reliability of the resulting annotations. We therefore also inspect the agreement on a per-question basis. Here Q2 (group size) reached the highest agreement at 0.772/0.781 (eval/test), while Q4 (physical activity) showed the lowest, yet still moderate agreement at 0.678/0.579.

The final ground truth was established using majority vote across the four annotators~\cite{snow2008cheap,sheng2008getAnotherLabel}. Due to the multi-label nature of the task, a label was selected when at least two annotators selected it (2-out-of-4 threshold). In evaluation, human annotators are scored against leave-one-out majority, excluding their own labels (3vs1), while LLM predictions will be evaluated against the full four-annotator consensus (4vs1).

In terms of F-score on the consensus annotation labels pooled across all four questions, the annotators achieved on average 78.6\% F-score (test, n=150) and 80.7\% F-score (evaluation, n=50) against the leave-one-out majority vote (Appendix Table~\ref{tab:iaa_combined}).

\subsection{LLM Annotation}

With minimal editing, the annotation guidelines were crafted into the LLM prompt (see Appendix~\ref{appendix-prompt} for the full prompt). We evaluated six state-of-the-art open-weight models to determine which could best replicate human performance. Additionally, using the best open-weight model, we investigated variants such as prompting all four questions simultaneously in one prompt, as compared to prompting each question individually in separate prompts, so as to establish whether the models benefit from holistic entity understanding versus focused attention on individual classification dimensions.


\subsection{Models}

We evaluated the following six state-of-the-art open-weight models:  \texttt{Qwen2.5-72B-Instruct}, a 72B dense instruction model trained on 18T tokens with RLHF supporting 131k-token context \cite{qwen25_2024}; \texttt{Qwen3-235B-A22B}, a 235B-parameter MoE (22B active) trained on 36T tokens with native chain-of-thought capabilities \cite{qwen3_2025}; \texttt{DeepSeek-R1-Distill-Llama-70B}, a 70B student model distilled from the 671B DeepSeek-R1 teacher with RL-driven chain-of-thought \cite{deepseek_r1_2025}; \texttt{Meta-Llama-3.3-70B-Instruct}, a December 2024 refresh with 70B parameters and 128k context \cite{llama3_2024}; \texttt{Llama-4-Scout-17B-16E}, a 17B-active/16-expert MoE enabling 10M-token contexts \cite{llama4_scout_2025}; \texttt{Mistral-Large-Instruct-2407}, a 123B flagship model with 128k context \cite{mistral_large_2024}. Notably, neither Qwen3 nor DeepSeek-R1, despite their chain-of-thought capabilities, outperformed Mistral-Large on this task. 

The overall results of the models in terms of F-score on the evaluation set (n=50) are summarized in Table~\ref{tab:model_comparison_avg}, demonstrating substantial differences between individual models.

\begin{table}[t]
\centering
\renewcommand{\arraystretch}{1.05}
\setlength{\tabcolsep}{8pt}

\begin{tabular}{lc}
\hline
\textbf{Model} & \textbf{Avg F1} \\
\hline
\texttt{Qwen2.5-72B-Instruct}              & 65.0 \\
\texttt{Qwen3-235B-A22B}                   & 68.2 \\
\texttt{DeepSeek-R1-Distill-Llama-70B}     & 62.5 \\
\texttt{Meta-Llama-3.3-70B-Instruct}         & 73.0 \\
\texttt{Llama-4-Scout-17B-16E}             & 70.4 \\
\texttt{Mistral-Large-Instruct-2407}       & \textbf{74.6} \\
\texttt{Mistral-Large-Instruct-2407}$^{\dagger}$ & 73.6 \\
\hline
\end{tabular}

\vspace{2mm}
\small $^{\dagger}$Same model, but prompted one question at a time rather than all four questions for each entity.
\caption{Average F-scores of each model on the evaluation set (n=50).}
\label{tab:model_comparison_avg}

\end{table}

\subsection{Prompt Optimization}

Among the six models evaluated, \texttt{Mistral-Large} achieved the highest performance with 74.6\% average F-score (Table~\ref{tab:model_comparison_avg}). To test whether automated prompt engineering could further improve this baseline, two optimizer models (OpenAI's o3 and Anthropic's Claude 4 Opus) were used to generate optimized prompt variants when being informed about the errors observed in the annotation carried out with the original prompt.

To this end, the 50-entity evaluation set was divided into 5 splits (20 items for optimization, 30 items for evaluation), with each entity appearing in exactly two optimization sets and three evaluation sets. For each split, both optimizer models produced two prompt variants based on error feedback, yielding 20 optimized prompts total. Only 7 showed improvements on the held-out data, and these gains were individually relatively minor, on average 4pp F-score (when averaging only data splits with an improvement, i.e.\ an overestimate of any actual gain). This suggests the iterative human annotation process had already produced instructions clear enough for the model to approximate human judgment.

\section{Results}

\begin{table}[th]
\centering
\small
\setlength{\tabcolsep}{4pt}
\begin{tabular}{llccccc}
\hline
\textbf{Method}  & \textbf{Schema} & \textbf{Q1} & \textbf{Q2} & \textbf{Q3} & \textbf{Q4} & \textbf{Avg} \\
\hline
Orig. prompt & Full & 77.0 & 85.4 & 73.6 & 54.6 & 72.7 \\
Best prompt & Full & 78.2 & 85.8 & 71.7 & 55.9 & 72.9 \\
\hline
7 prompt ens. & Coarse & 80.0 & 87.7 & 76.8 & 62.7 & 76.8 \\
7× orig. ens. & Coarse & 76.9 & 87.4 & 79.9 & 60.0 & 76.1 \\
\hline
7 prompt ens. & Full & 80.0 & 85.0 & 71.4 & 58.9 & 73.7 \\
7× orig. ens. & Full & 76.9 & 85.9 & 75.3 & 55.0 & 73.3 \\
\hline
Human & Full & 82.5 & 88.6 & 73.8 & 69.3 & 78.6 \\
Human & Coarse & 82.5 & 90.1 & 81.2 & 73.4 & 81.8 \\
\hline
\end{tabular}
\caption{Performance of single-prompt methods, ensemble methods, and human annotators on test set (n=150) in full and simplified coarse label schema. All values are F-score.}
\label{tab:simplified_labels}

\end{table}

The entire framework was evaluated against the final ground truth test set of 150 entities, split evenly between hobbies and organizations. The results are summarized in Table~\ref{tab:simplified_labels}, demonstrating that human annotators clearly outperform the LLM (72.7\% vs.\ 78.6\% F-score), however the difference is not even across the four annotation questions. From the 20 automatically-optimized prompts, only one outperformed the original human-engineered prompt, achieving 72.9\% F-score compared to 72.7\% F-score for the original, a marginal gain of 0.2 percentage points. 

We further evaluated two ensemble approaches: a majority vote across the 7 optimized prompts which have previously shown an improvement on the evaluation set, and for comparison a majority vote across 7 runs of the original prompt (leveraging the fact that the LLMs sample in generation, and the output therefore is not deterministic). As seen in Table~\ref{tab:simplified_labels}, the 7-prompt ensemble achieved 73.7\% F-score (93.8\% of human baseline), compared to 73.3\% for the repeated original prompt. Simply running the same prompt multiple times showed minor improvement (73.3\% vs 72.7\%), suggesting that ensemble voting can partially compensate for the stochastic nature of LLM outputs.
The most notable improvement came on Q1 (Categories), where the optimized ensemble achieved 80.0\% F-score versus the repeated original's 76.9\%. However, the optimized ensemble performed worse on Q3 (Frequency), suggesting that different prompts may have complementary strengths and weaknesses.

\subsection{Coarse Label Schema: Improving Performance Through Reduced Granularity}

Considering that the downstream use of the labeled data requires a maximal attainable accuracy, we also experimented with reduced label granularity, producing a coarse schema by treating the following labels as equivalent: \textbf{Q2:} ``Small group'' = ``Large group''; \textbf{Q3:} ``Occasional'' = ``Event-based''; \textbf{Q4:} ``Intense'' = ``Continuous'' = ``Light''. This simplification essentially reduces Q2 to the choice of ``solitary vs social,'' Q3 to the choice of ``regular vs rare,'' and Q4 to the choice ``active vs stationary,'' while keeping Q1's full categorical granularity. 

With these coarse labels, the optimized ensemble achieved 76.8\% F1 score, representing 93.9\% of human baseline. This represents a substantial 3.1 percentage point improvement over the same ensemble with full label granularity (73.7\%). The gains were most pronounced in Q2 (Group Size), where collapsing ``Small group'' and ``Large group'' into a single ``Social'' category improved performance from 85.0\% to 87.7\%.

\subsection{Processing the Complete Dataset}

Based on the results outlined above, the full dataset of 60,647 entities was processed with the 7-prompt ensemble method, resulting in total of 424,529 LLM queries. The run was carried out on a cluster system using the vLLM library for efficient LLM generation. The system achieved 2219.38 tokens/s input and 54.05 tokens/s output. Despite identical formatting instructions, the prompts varied in reliability: the best required only 1.1 attempts per entity on average, while the most problematic needed 1.9 attempts (maximum 5 attempts per entity). Most failures stemmed from malformed JSON or empty responses; others involved incorrect category choices or hallucinated words. After majority voting across all 7 prompts, we obtained the final LLM annotation of all entities.

For Q1 (Categories), the most common labels were Administrative (19.5\%, n=11,844), Professional/Work-related (17.9\%, n=10,830), and General Social Group (12.8\%, n=7,744), reflecting the formal organizational structure of post-war Finnish society. Multi-label classification proved essential for Q1: 29.2\% of entities (17,717) received multiple category labels, with all 18 categories appearing in multi-label combinations (Table~\ref{tab:multilabel_summary}). Administrative entities most frequently combined with Professional/Work-related (1,654 cases), Educational/Academic (1,583), and Social Welfare (1,072). In contrast, Q2--Q4 showed minimal multi-label usage (1.8--6.6\%), validating our design where Q1 captures overlapping domains while Q2--Q4 capture more singular dimensions.
The full distribution across all labels is listed in Appendix \ref{appendix-full-distribution-unique}.

\begin{table}[h]
\centering
\caption{Multi-label entity distribution across questions.}
\label{tab:multilabel_summary}
\small
\begin{tabular}{lrrr}
\hline
\textbf{Question} & \textbf{Multi-label \%} & \textbf{Total} & \textbf{Labels} \\
\hline
Q1 – Category & 29.2\% & 17,717 & 18 \\
Q2 – Group size & 6.6\% & 4,014 & 3 \\
Q3 – Frequency & 1.8\% & 1,076 & 4 \\
Q4 – Physical act. & 2.5\% & 1,511 & 4 \\
\hline
\end{tabular}
\end{table}

\subsubsection{Mention-Weighted Analysis}

All results so far were reported for unique entities. Weighting by mention frequency of each unique entity reveals distinct patterns between hobbies and organizations (full details are in Appendix \ref{appendix-full-distribution-weighted}). Hobbies concentrate heavily in Creative/Artistic (36.3\%) and Nature-related (32.9\%) categories, while organizations distribute across General Social Group (19.0\%), Administrative (12.6\%), and Professional/Work-related (12.3\%). These choices represent the final dataset and label distribution across the data. Group size (Q2) shows an interesting contrast whereby 96.1\% of organizational mentions involve large groups, while hobbies split between solitary (43.4\%) and small group (40.7\%) activities. For frequency (Q3), 79.5\% of organizational mentions are occasional, reflecting periodic meetings, whereas 66.0\% of hobby mentions are continuous activities. Physical activity (Q4) distributions also diverge: 68.2\% of organizations are stationary versus only 22.4\% of hobbies, with most hobbies involving light (55.5\%) or intense (10.0\%) physical activity.
These patterns validate our multi-dimensional approach and provide quantitative characterization of social participation patterns in the Karelian refugee population suitable for downstream health outcomes research.

\subsection{Dimension Reliability by Category (Q1)}

Any downstream analysis will need to take into account the reliability of both human and LLM predictions. For instance, Q4 (Physical Activity) performed clearly worst on the test set (Table \ref{tab:simplified_labels}). To investigate whether the prediction reliability depends on the entity's category type, we analyzed Q2--Q4 reliability conditional on the answer to Q1. This addresses a question for downstream research: given that an entity is classified as a certain category (Q1), how reliable are answers to the other dimensions Q2--Q4? The statistics are shown in  Table~\ref{tab:reliability_by_q1_full}.

\begin{table}[t]
\centering
\small
\setlength{\tabcolsep}{3pt}
\begin{tabular}{lrrrrr}
\hline
\textbf{Q1 Category} & \textbf{n} & \textbf{Q1} & \textbf{Q2} & \textbf{Q3} & \textbf{Q4} \\
\hline
Creative/Artistic & 33 & \colorbox{green!25}{93/95} & \colorbox{green!25}{85/91} & \colorbox{green!25}{90/94} & \colorbox{yellow!40}{51/87} \\
Professional & 27 & \colorbox{green!25}{77/88} & \colorbox{green!25}{91/96} & \colorbox{yellow!40}{70/81} & \colorbox{yellow!40}{63/77} \\
Sports/Physical & 24 & \colorbox{green!25}{82/93} & \colorbox{green!25}{81/89} & \colorbox{yellow!40}{72/85} & \colorbox{green!25}{76/90} \\
Administrative & 21 & \colorbox{green!25}{85/88} & \colorbox{green!25}{95/96} & \colorbox{green!25}{76/79} & \colorbox{green!25}{84/83} \\
Nature-related & 19 & \colorbox{green!25}{79/92} & \colorbox{green!25}{75/86} & \colorbox{yellow!40}{60/71} & \colorbox{yellow!40}{58/72} \\
General social & 12 & \colorbox{green!25}{75/88} & \colorbox{green!25}{92/97} & \colorbox{red!25}{47/82} & \colorbox{red!25}{0/65} \\
Educational & 10 & \colorbox{yellow!40}{67/86} & \colorbox{green!25}{90/96} & \colorbox{yellow!40}{73/82} & \colorbox{green!25}{90/95} \\
Property/Own. & 10 & \colorbox{yellow!40}{67/69} & \colorbox{yellow!40}{73/82} & \colorbox{yellow!40}{63/85} & \colorbox{yellow!40}{80/72} \\
Animal-related & 7 & \colorbox{green!25}{83/94} & \colorbox{green!25}{79/86} & \colorbox{green!25}{81/82} & \colorbox{yellow!40}{71/73} \\
Special hobbies & 7 & \colorbox{yellow!40}{50/84} & \colorbox{green!25}{90/91} & \colorbox{green!25}{90/87} & \colorbox{yellow!40}{67/76} \\
Military-related & 6 & \colorbox{green!25}{91/91} & \colorbox{green!25}{83/92} & \colorbox{red!25}{44/78} & \colorbox{red!25}{0/79} \\
Religious/Spirit. & 6 & \colorbox{green!25}{80/92} & \colorbox{green!25}{86/99} & \colorbox{yellow!40}{67/87} & \colorbox{yellow!40}{67/78} \\
Social welfare & 5 & \colorbox{yellow!40}{59/67} & \colorbox{green!25}{100/100} & \colorbox{red!25}{40/76} & \colorbox{red!25}{20/78} \\
\hline
\end{tabular}

\caption{Q1--Q4 reliability conditioned by entity category in Q1, showing both model and human agreement rates. Format: Model\% / Human\%. Colors: \colorbox{green!25}{$\geq$75\%}, \colorbox{yellow!40}{50--74\%}, \colorbox{red!25}{$<$50\%} (based on lower score).}
\label{tab:reliability_by_q1_full}

\end{table}

For instance: Sports/Physical activity entities (n=24), the model correctly identified the Q1 category 82\% of the time. Among these sports-related entities, Q2 (Group Size) achieved 81\% accuracy, Q3 (Frequency) 72\%, and Q4 (Physical Activity) 76\%. In contrast Creative/Artistic entities (n=33) achieved 93\% Q1 accuracy but only 51\% on Q4, reflecting difficulty distinguishing stationary activities  from movement-based ones.

\section{Error analysis}

As shown in Table~\ref{tab:model-human-disagreement}, 63\% of entities show perfect agreement or low disagreement between the model and human annotators. The full listing of entities and their human and model agreement levels is available in the Appendix~\ref{appendix-full-entity-listing}. We carry out a manual inspection of several main classes of entities by agreement: 
\begin{table}
\centering
\begin{tabular}{|l|r|}
\hline
Perfect agreement (0\%) & 19\%\\
Low disag. (<25\%) & 44\%\\
Moderate disag. (25--50\%) & 30\%\\
High disag. (>50\%) & 7\%\\
\hline
\end{tabular}
\caption{Degrees of model--human disagreement.}
\label{tab:model-human-disagreement}
\end{table}

\paragraph{Perfect agreement:} Organizations with clear categories, group sizes, and physical activity levels are straightforward cases for the model. Examples: Choir (\emph{Laulukuoro}), Bandy (\emph{jääpallo}), Basketball and Volleyball (\emph{kori- ja lentopalloilu}), Biathlon (\emph{ampumahiihto}).

\paragraph{Model blind spots:} Entities showing a high model disagreement despite high degree of human agreement. Typical cases are abbreviations without explicit context, and culturally-specific items such as rare Finnish folk dances. These are often known to humans familiar with Finnish culture and history, but unknown to general open-weight LLMs. Another typical case is the model's overconfidence, making a guess, where humans resort to the ``Cannot be determined'' label. Typical examples are e.g.\ Seniors' circle (\emph{vanhusten piiri}), and Lotta Svärd activities (\emph{lottatoiminta}).

\paragraph{Genuine Ambiguity:} Entities with high both model and human disagreement include e.g.\ Motor sports (\emph{moottoriurheilu}), where the human annotators did not agree on frequency, group size, nor degree of physical activity. Another typical class are abbreviations, such as the military regiment abbreviation (\emph{2.KKK/KKR}) and entities with OCR errors, where some annotators attempted disambiguation, while others marked a ``Data error''.

\section{Conclusions}

We demonstrate that LLMs can implement social organization and leisure activity classification according to a custom taxonomy developed by domain experts for a specific analytic task. The models achieve 76.8\% F-score, which is 93.9\% of human baseline performance on coarse labels, and 93.8\% on full granularity. This includes the added complexity of a relatively small language (Finnish) and the need of multi-label classification, where we see the LLMs are capable of assigning multiple labels, allowing for a structured multi-question schema, where some questions (e.g. Q1) are inherently multi-label.

The performance of the LLM at nearly 94\% of human annotation provides a worthwhile alternative to manual classification of tens of thousands of entities, especially if any downstream analysis can be designed to take into account the levels of uncertainty necessarily present in such a categorization task.

To this end, we find that analyzing and quantifying LLM classification errors is crucial, since these errors are rarely random. They follow consistent patterns: some reflect genuine failures of model understanding, while others represent interpretive differences where the model's reasoning diverges from human consensus rather than being outright wrong. Characterizing these error types, identifying which options are systematically misclassified, and providing reliability measures for specific categories is essential for downstream use of this data.

All data will be made available upon publication under an open license.

\section{Acknowledgments}

This work was carried out in the Human Diversity University profilation programme (PROFI-7) of the Research Council of Finland (grant 352727), and in part supported also through the KinSocieties, ERC-2022-ADG grant number 101098266. Computational resources were provided by CSC — the Finnish IT Center for Science.

\section{Limitations}

LLMs show consistent failure patterns, including uncertainty avoidance and cultural knowledge gaps. Rather than selecting ``Cannot be determined,'' the models tend to guess concrete answers even when annotators acknowledge ambiguity. Similarly, Finnish-specific terminology and regional abbreviations lead to systematic misclassification.

Additionally developing a custom taxonomy requires substantial annotation effort. Our annotation framework involved five rounds of refinement with four annotators before achieving stable guidelines. Comparing this to established schemes like NER, where pretrained models and labeled datasets exist, this represents a significant overhead. Furthermore, if reliability analysis reveals poor alignment for specific categories, additional work is required, either more annotations to improve agreement, or refining the categories themselves. However while the framework is designed to address the specific research questions related to our data and may not be directly transferable to others, we believe that many of its categories are of universal interest (e.g. in health research), and that the framework can serve as a reference to be adapted for other works describing social participation.


\bibliography{custom}

\clearpage
\onecolumn
\appendix\section{Appendix}

\subsection{Task prompt}
\label{appendix-prompt}
\begin{Verbatim}[breaklines=true,breakanywhere=true,fontsize=\small]
### TASK

Classify the entity given below by answering four questions (q1–q4).  
For each question, choose the option(s) that **most accurately** match the entity, following the instructions provided.  
Write your answers exactly according to the **CORRECT ANSWER FORMAT** structure given below.

---


---

## QUESTIONS AND OPTIONS

**Always clearly separate the option and its description, e.g.**  
`Professional/Work-related – Trade unions, professional associations, workplace-related groups…`

### Question 1 (q1) – Categories (select **all** that apply)

| Option | Description |
|--------|-------------|
| Professional/Work-related | Trade unions, professional associations, workplace-related groups; |
| Cultural/Traditional | Cultural heritage, local culture, traditions (e.g. Karelian Society) |
| Religious/Spiritual | Church activities, religious education, spiritual associations |
| Sports/Physical activity | All kinds of sports, exercise, team sports and individual sports |
| Creative/Artistic | Music, visual arts, handicrafts, theatre, dance, etc. |
| Educational/Academic | Studying, education, scholarly associations, study circles |
| Social welfare | Charity work, volunteering, community service |
| Administrative | Public administration roles and bodies (municipality, church, state), non-political |
| Political | Parties, party organizations, political advocacy |
| General social group | Martta associations, youth clubs, general-purpose community groups |
| Nature-related | Hunting, fishing, hiking, nature conservation |
| Health-related | Peer-support groups, support for people with illnesses (e.g. Rheumatism Association) |
| Property/Ownership | Road maintenance associations, housing companies, detached-house maintenance associations |
| Non-physical games | Chess, board games, role-playing and video games |
| Animal care/Hobby | Beekeeping, horse riding, dog shows |
| Special hobbies | Stamp collecting, radio technology, etc. |
| Cooking | Food preparation, baking |
| Military-related | National defence, reservist activities |
| Not definable | Impossible to determine from name/context |
| Data error | Entity is not an organization/hobby OR is a data error |

---

### Question 2 (q2) – Group size (select the most typical one(s))

| Option | Description |
|--------|-------------|
| Alone | 1 person; the activity can be done completely independently |
| Small group | 2–5 people; close cooperation between participants |
| Large group | > 6 people; requires broader organization and structures |
| Not definable | Group size varies / cannot be inferred |
| Data error | Data issue prevents assessment |

---

### Question 3 (q3) – Frequency of activity (select the most likely one(s))

| Option | Description |
|--------|-------------|
| Regular | Fixed schedule, e.g. weekly |
| Occasional | About once a month, not fully fixed |
| Event-based | Once a year or less often, in connection with an event |
| Continuous | Ongoing activity without fixed schedule; can be done any time |
| Not definable | Frequency is unclear |
| Data error | Data issue prevents assessment |

---

### Question 4 (q4) – Level of movement / Physical activity (select the most likely one(s))

| Option | Description |
|--------|-------------|
| Intense | Heavy sports, intensive muscular work |
| Continuous | Long-lasting steady movement (e.g. hiking) |
| Light | Occasional light movement, light activity/puttering |
| Stationary | Minimal movement, sitting/standing (meetings, administration) |
| Not definable | Physical activity level cannot be determined |
| Data error | Data issue prevents assessment |

---

## ANNOTATION GUIDELINES

### General principles
- Accuracy > coverage → if uncertain, choose **“Not definable”**.  
- If the entity is **not** an organization/hobby, use **“Data error”** for all questions.  
- Base decisions on the available information; avoid guessing.
- If provided, take into account the hierarchy in which the organization is mentioned. If it is empty or UNK, you can ignore it. 

### Question-specific notes  
**q1:** Select at least one category; some entities belong to multiple categories.  
* Administrative is not Political: administrative = official governing bodies, political = parties.

**q2:** Choose the most typical group size; meetings/organizations → often **Large group**.  
* If the estimate is difficult → **Not definable**.

**q3:**  
* **Continuous** = independent, no group required.  
* **Regular** = active weekly/monthly meetings.  
* **Occasional** = committees & boards (~ once/month).  
* **Event-based** = rarer, 1–2 times/year.  
* Unclear → **Not definable**.

**q4:**  
* Administrative body → **Stationary**; sports club → **Intense**.  
* Board of a sports club → **Stationary**.  
* Unclear → **Not definable**.

### EXAMPLES  

|--------|----|----|----|----|
| football | Sports/Physical activity | Large group | Regular | Intense |
| rheumatism association | Health-related | Large group | Regular | Stationary |
| mushroom picking | Nature-related | Alone, Small group | Continuous | Light |
| social welfare committee | Administrative, Social welfare | Large group | Occasional | Stationary |
| Guards band | Military-related, Creative/Artistic | Large group | Regular | Stationary |
| Järvenpää youth board | Administrative | Large group | Occasional | Stationary |
| housekeeping | Property/Ownership | Alone, Small group | Continuous | Light |
| local party branch | Political | Large group | Occasional | Stationary |

---  
#### CORRECT ANSWER FORMAT  

[Answer begin]
{{
  "{{entity_name}}": {{
    "q1": [...],
    "q2": [...],
    "q3": [...],
    "q4": [...]
  }}
}}
[Answer end]

#### Entity to be annotated  

"{entity_name}"  

Hierarchy: "{hierarchies}"

"{past_mistakes}"


\end{Verbatim}

\subsection{Inference Hyperparameters}
\label{appendix-hyperparameters}
Inference was performed on the LUMI supercomputer using vLLM as the inference engine with the following sampling parameters: \texttt{temperature=0.3}, \texttt{top\_p=1.0}, \texttt{top\_k=40}, \texttt{max\_tokens=300}. Processing used batch size of 150.

\subsection{Inter-annotator agreement details}
\begin{table*}[h]
\centering
\small
\setlength{\tabcolsep}{5pt}
\begin{tabular}{lcccccc}
\hline
\textbf{Annotator Pair} & \textbf{Q1} & \textbf{Q2} & \textbf{Q3} & \textbf{Q4} & \textbf{Average} \\
\hline
\multicolumn{6}{c}{\textbf{Pairwise Cohen's Kappa -- Evaluation Set (n=50)}} \\
\hline
Annotator 1 vs 2 & 0.800 & 0.806 & 0.667 & 0.721 & 0.749 \\
Annotator 1 vs 3 & 0.753 & 0.743 & 0.725 & 0.711 & 0.733 \\
Annotator 1 vs 4 & 0.753 & 0.767 & 0.524 & 0.660 & 0.676 \\
Annotator 2 vs 3 & 0.684 & 0.753 & 0.677 & 0.688 & 0.701 \\
Annotator 2 vs 4 & 0.697 & 0.848 & 0.651 & 0.754 & 0.737 \\
Annotator 3 vs 4 & 0.683 & 0.714 & 0.571 & 0.535 & 0.626 \\
\textbf{Mean} & \textbf{0.728} & \textbf{0.772} & \textbf{0.636} & \textbf{0.678} & \textbf{0.704} \\
\hline
\multicolumn{6}{c}{\textbf{Pairwise Cohen's Kappa -- Test Set (n=150)}} \\
\hline
Annotator 1 vs 2 & 0.783 & 0.825 & 0.642 & 0.620 & 0.718 \\
Annotator 1 vs 3 & 0.824 & 0.790 & 0.698 & 0.528 & 0.710 \\
Annotator 1 vs 4 & 0.778 & 0.776 & 0.581 & 0.723 & 0.714 \\
Annotator 2 vs 3 & 0.737 & 0.751 & 0.636 & 0.507 & 0.658 \\
Annotator 2 vs 4 & 0.712 & 0.812 & 0.649 & 0.680 & 0.713 \\
Annotator 3 vs 4 & 0.767 & 0.732 & 0.602 & 0.415 & 0.629 \\
\textbf{Mean} & \textbf{0.767} & \textbf{0.781} & \textbf{0.634} & \textbf{0.579} & \textbf{0.690} \\
\hline
\multicolumn{6}{c}{\textbf{Individual F1 vs Leave-One-Out Majority -- Evaluation Set (n=50)}} \\
\hline
Annotator 1 & 0.867 & 0.873 & 0.813 & 0.807 & 0.840 \\
Annotator 2 & 0.793 & 0.907 & 0.833 & 0.793 & 0.832 \\
Annotator 3 & 0.756 & 0.813 & 0.833 & 0.733 & 0.784 \\
Annotator 4 & 0.793 & 0.880 & 0.710 & 0.713 & 0.774 \\
\textbf{Mean} & \textbf{0.802} & \textbf{0.868} & \textbf{0.797} & \textbf{0.762} & \textbf{0.807} \\
\hline
\multicolumn{6}{c}{\textbf{Individual F1 vs Leave-One-Out Majority -- Test Set (n=150)}} \\
\hline
Annotator 1 & 0.865 & 0.898 & 0.749 & 0.744 & 0.814 \\
Annotator 2 & 0.800 & 0.902 & 0.756 & 0.722 & 0.795 \\
Annotator 3 & 0.830 & 0.869 & 0.753 & 0.580 & 0.758 \\
Annotator 4 & 0.805 & 0.876 & 0.696 & 0.724 & 0.775 \\
\textbf{Mean} & \textbf{0.825} & \textbf{0.886} & \textbf{0.738} & \textbf{0.693} & \textbf{0.786} \\
\hline
\end{tabular}
\vspace{1mm}
\par\noindent\footnotesize{Note: Krippendorff's Alpha values are nearly identical to Cohen's Kappa (within 0.001) and omitted for brevity.}
\caption{Inter-annotator agreement metrics across evaluation (n=50) and test (n=150) sets.}
\label{tab:iaa_combined}

\end{table*}

\clearpage
\subsection{Full distribution of unique entities}
\label{appendix-full-distribution-unique}
\begin{table*}[htbp]
\centering
\caption{Distribution across 60,647 unique entities for all four questions.}
\label{tab:distribution_combined}
\small
\begin{tabular}{lrr|lrr}
\hline
\multicolumn{3}{c|}{\textbf{Q1 -- Category}} & \multicolumn{3}{c}{\textbf{Q2 -- Group Size}} \\
\textbf{Label} & \textbf{n} & \textbf{\%} & \textbf{Label} & \textbf{n} & \textbf{\%} \\
\hline
Administrative & 11,844 & 19.5 & Large group & 53,574 & 88.3 \\
Professional & 10,830 & 17.9 & Small group & 4,889 & 8.1 \\
General Social & 7,744 & 12.8 & Alone & 4,840 & 8.0 \\
Creative & 7,396 & 12.2 & Data error & 853 & 1.4 \\
Sports & 7,328 & 12.1 & Not definable & 654 & 1.1 \\
Social Welfare & 5,760 & 9.5 & & & \\
Educational & 4,921 & 8.1 & & & \\
Cultural & 3,846 & 6.3 & & & \\
Military & 3,798 & 6.3 & & & \\
Religious & 3,166 & 5.2 & & & \\
Property & 2,696 & 4.4 & & & \\
Nature & 2,590 & 4.3 & & & \\
Political & 2,034 & 3.4 & & & \\
Health & 1,930 & 3.2 & & & \\
Special hobbies & 958 & 1.6 & & & \\
Animal care & 886 & 1.5 & & & \\
Data error & 853 & 1.4 & & & \\
Not definable & 283 & 0.5 & & & \\
Cooking & 176 & 0.3 & & & \\
Non-phys. games & 175 & 0.3 & & & \\
\hline
\multicolumn{3}{c|}{\textbf{Q3 -- Frequency}} & \multicolumn{3}{c}{\textbf{Q4 -- Physical}} \\
\textbf{Label} & \textbf{n} & \textbf{\%} & \textbf{Label} & \textbf{n} & \textbf{\%} \\
\hline
Occasional & 42,999 & 70.9 & Stationary & 41,103 & 67.8 \\
Regular & 11,800 & 19.5 & Light & 9,902 & 16.3 \\
Continuous & 5,092 & 8.4 & Intense & 6,291 & 10.4 \\
Data error & 853 & 1.4 & Not definable & 2,567 & 4.2 \\
Event-based & 555 & 0.9 & Continuous & 1,542 & 2.5 \\
Not definable & 490 & 0.8 & Data error & 853 & 1.4 \\
\hline
\end{tabular}
\end{table*}

\clearpage
\subsection{Full distribution of entity occurrences}
\label{appendix-full-distribution-weighted}

\begin{table*}[h!]
\centering

\begin{tabular}{lrrrrrr}
\hline
& \multicolumn{2}{c}{\textbf{Hobbies}} & \multicolumn{2}{c}{\textbf{Organizations}} & \multicolumn{2}{c}{\textbf{Total}} \\
\textbf{Label} & \textbf{Count} & \textbf{\%} & \textbf{Count} & \textbf{\%} & \textbf{Count} & \textbf{\%} \\
\hline
\multicolumn{7}{c}{\textbf{Q2 -- Group Size}} \\
\hline
Large group & 55,252 & 12.9 & 198,318 & 96.1 & 253,570 & 39.9 \\
Alone & 186,679 & 43.4 & 512 & 0.2 & 187,191 & 29.4 \\
Small group & 174,877 & 40.7 & 4,104 & 2.0 & 178,981 & 28.1 \\
Not definable & 12,232 & 2.8 & 1,140 & 0.6 & 13,372 & 2.1 \\
Data error & 687 & 0.2 & 2,387 & 1.2 & 3,074 & 0.5 \\
\hline
\multicolumn{7}{c}{\textbf{Q3 -- Frequency}} \\
\hline
Occasional & 43,018 & 16.0 & 163,126 & 79.5 & 206,144 & 43.4 \\
Continuous & 177,956 & 66.0 & 1,804 & 0.9 & 179,760 & 37.8 \\
Regular & 41,123 & 15.2 & 36,439 & 17.8 & 77,562 & 16.3 \\
Not definable & 6,025 & 2.2 & 1,060 & 0.5 & 7,085 & 1.5 \\
Data error & 687 & 0.3 & 2,387 & 1.2 & 3,074 & 0.6 \\
Event-based & 865 & 0.3 & 448 & 0.2 & 1,313 & 0.3 \\
\hline
\multicolumn{7}{c}{\textbf{Q4 -- Physical Activity}} \\
\hline
Stationary & 59,508 & 22.4 & 145,285 & 68.2 & 204,793 & 42.7 \\
Light & 147,726 & 55.5 & 31,210 & 14.6 & 178,936 & 37.3 \\
Intense & 26,605 & 10.0 & 16,138 & 7.6 & 42,743 & 8.9 \\
Not definable & 11,513 & 4.3 & 14,497 & 6.8 & 26,010 & 5.4 \\
Continuous & 19,898 & 7.5 & 3,663 & 1.7 & 23,561 & 4.9 \\
Data error & 687 & 0.3 & 2,387 & 1.1 & 3,074 & 0.6 \\
\hline
\end{tabular}
\caption{Mention-weighted distribution (355,648 total mentions) for Q2--Q4 by entity type.}
\label{tab:weighted_combined}

\end{table*}


\begin{table*}[h!]
\centering

\begin{tabular}{lrrrrrr}
\hline
& \multicolumn{2}{c}{\textbf{Hobbies}} & \multicolumn{2}{c}{\textbf{Organizations}} & \multicolumn{2}{c}{\textbf{Total}} \\
\textbf{Category} & \textbf{Count} & \textbf{\%} & \textbf{Count} & \textbf{\%} & \textbf{Count} & \textbf{\%} \\
\hline
Creative/Artistic & 97,180 & 36.3 & 16,639 & 6.3 & 113,819 & 21.4 \\
Nature-related & 88,174 & 32.9 & 3,880 & 1.5 & 92,054 & 17.3 \\
Sports/Physical & 46,269 & 17.3 & 15,390 & 5.8 & 61,659 & 11.6 \\
General Social Group & 1,621 & 0.6 & 50,089 & 19.0 & 51,710 & 9.7 \\
Professional/Work & 1,811 & 0.7 & 32,344 & 12.3 & 34,155 & 6.4 \\
Administrative & 235 & 0.1 & 33,301 & 12.6 & 33,536 & 6.3 \\
Cultural/Traditional & 565 & 0.2 & 31,685 & 12.0 & 32,250 & 6.1 \\
Educational/Academic & 14,372 & 5.4 & 12,112 & 4.6 & 26,484 & 5.0 \\
Social Welfare & 1,055 & 0.4 & 21,406 & 8.1 & 22,461 & 4.2 \\
Religious/Spiritual & 2,678 & 1.0 & 12,440 & 4.7 & 15,118 & 2.8 \\
Military & 433 & 0.2 & 13,858 & 5.3 & 14,291 & 2.7 \\
Property/Ownership & 3,931 & 1.5 & 5,617 & 2.1 & 9,548 & 1.8 \\
Health-related & 136 & 0.1 & 5,482 & 2.1 & 5,618 & 1.1 \\
Animal care/hobby & 3,689 & 1.4 & 967 & 0.4 & 4,656 & 0.9 \\
Political & 220 & 0.1 & 4,216 & 1.6 & 4,436 & 0.8 \\
Special hobbies & 3,239 & 1.2 & 799 & 0.3 & 4,038 & 0.8 \\
Data error & 687 & 0.3 & 2,387 & 0.9 & 3,074 & 0.6 \\
Not definable & 278 & 0.1 & 918 & 0.3 & 1,196 & 0.2 \\
Non-physical games & 880 & 0.3 & 211 & 0.1 & 1,091 & 0.2 \\
Cooking & 565 & 0.2 & 28 & 0.0 & 593 & 0.1 \\
\hline
\end{tabular}
\caption{Q1 category distribution weighted by mention frequency (355,648 total mentions).}
\label{tab:q1_weighted}

\end{table*}

\clearpage
\subsection{Full test set categorization}
\label{appendix-full-entity-listing}
\scriptsize
\setlength{\tabcolsep}{4pt}
\begin{longtable}{r p{7cm} c c}
\textbf{Rank} & \textbf{Entity} & \textbf{Model Disagr} & \textbf{Human Disagr} \\
\hline
\endfirsthead
\textbf{Rank} & \textbf{Entity} & \textbf{Model Disagr} & \textbf{Human Disagr} \\
\hline
\endhead
1 & Loimaa Co-op Bank (\textit{Loimaan KOP}) & \textbf{100.0\%} & 66.7\% \\
2 & Folk dance (\textit{tanhu}) & \textbf{100.0\%} & 50.7\% \\
3 & Poor relief fund (\textit{Vaivaiskassa}) & \textbf{100.0\%} & 66.7\% \\
4 & Military regiment abbr. (\textit{2.KKK/KKR}) & 91.7\% & 75.0\% \\
5 & Association abbr. (\textit{KTV}) & \textbf{87.5\%} & 25.0\% \\
6 & Summer cottage life (\textit{kesämökkielämä}) & \textbf{79.2\%} & 56.2\% \\
7 & Sea excursions (\textit{retkeily merellä}) & \textbf{79.2\%} & 54.2\% \\
8 & Boating sports (\textit{veneurheilu}) & 79.2\% & 65.3\% \\
9 & Nokia SOK evening club (\textit{Nokian SOK:n illkkakerho}) & \textbf{75.0\%} & 50.0\% \\
10 & Council (\textit{valtuusto}) & 75.0\% & 68.8\% \\
11 & Seniors' circle (\textit{vanhusten piiri}) & \textbf{75.0\%} & 18.8\% \\
12 & Self-employed entrepreneurs (\textit{Yksityisyrittäjät}) & \textbf{75.0\%} & 20.8\% \\
13 & Ylöjärvi TPSL branch (\textit{Ylöjärven TPSL}) & \textbf{75.0\%} & 27.8\% \\
14 & Staying at summer cottage (\textit{kesämökillä oleskelu}) & 66.7\% & 52.1\% \\
15 & Summer cottage stays (\textit{kesämökkeily}) & 66.7\% & 52.1\% \\
16 & Motor sports (\textit{moottoriurheilu}) & 66.7\% & 75.0\% \\
17 & Camping/tents (\textit{telttailu}) & \textbf{66.7\%} & 25.0\% \\
18 & Karkkula Farmers' Association (\textit{Karkkulan Maamiesseura}) & \textbf{62.5\%} & 31.9\% \\
19 & Bird banding (\textit{lintujen merkitsijänä toimiminen}) & \textbf{62.5\%} & 41.7\% \\
20 & Martha tasks (\textit{martta-työt}) & \textbf{62.5\%} & 29.2\% \\
21 & Martha organization work (\textit{marttatyö}) & \textbf{62.5\%} & 29.2\% \\
22 & Attending sewing circles (\textit{ompeluseuroissa käyminen}) & \textbf{54.2\%} & 31.2\% \\
23 & War veterans (\textit{Sotaveteraanit}) & \textbf{54.2\%} & 27.1\% \\
24 & Hauho Farmers' Association (\textit{Hauhon Maamiesseura}) & \textbf{50.0\%} & 29.9\% \\
25 & Hämeenlinna Camera Club (\textit{Hämeenlinnan Kameraseura}) & \textbf{50.0\%} & 29.2\% \\
26 & Dev. disabilities support (\textit{Kehitysvammaisten tukiyhdistys}) & \textbf{50.0\%} & 29.9\% \\
27 & Lotta Svärd organization (\textit{Lotta-Svärd-järjestö}) & \textbf{50.0\%} & 20.8\% \\
28 & Lotta Svärd activities (\textit{lottatoiminta}) & \textbf{50.0\%} & 20.8\% \\
29 & Martha activities (\textit{marttatoiminta}) & \textbf{50.0\%} & 25.0\% \\
30 & Travels (\textit{matkustelee}) & 50.0\% & 66.7\% \\
31 & Opera (\textit{ooppera}) & 50.0\% & 56.2\% \\
32 & Local horse association (\textit{paikallinen hevosyhdistys}) & 50.0\% & 52.8\% \\
33 & Gliding (\textit{purjelento}) & 50.0\% & 64.6\% \\
34 & Chess (\textit{shakki}) & \textbf{50.0\%} & 20.8\% \\
35 & Betting (\textit{veikkaus}) & 50.0\% & 58.3\% \\
36 & Diaconia Committee (\textit{Diakoniatoimikunta}) & 45.8\% & 34.7\% \\
37 & Workers' association (\textit{työväenyhdistys}) & 45.8\% & 41.7\% \\
38 & Church attendance (\textit{kirkossakäyminen}) & \textbf{41.7\%} & 0.0\% \\
39 & East Karelia Folk College (\textit{Itä-Karjalan kansanopisto}) & 37.5\% & 33.3\% \\
40 & Making canal boards (\textit{kanavataulujen teko}) & \textbf{37.5\%} & 0.0\% \\
41 & Middle-distance running (\textit{keskipitkien matkojen juoksu}) & \textbf{37.5\%} & 0.0\% \\
42 & Fitness gymnastics (\textit{kuntovoimistelu}) & 37.5\% & 33.3\% \\
43 & Handicrafts (\textit{käsityö}) & \textbf{37.5\%} & 6.2\% \\
44 & Agricultural Producers (\textit{Maataloustuottajat}) & 37.5\% & 27.1\% \\
45 & Martha Association (\textit{Martta-yhdistys}) & 37.5\% & 27.1\% \\
46 & Mosaic work (\textit{mosaiikkityöt}) & \textbf{37.5\%} & 6.2\% \\
47 & Archaeological artifacts (\textit{muinaistieteellisten esineiden kerääminen}) & 37.5\% & 45.8\% \\
48 & Drawing (\textit{piirustus}) & \textbf{37.5\%} & 0.0\% \\
49 & Ice fishing (\textit{pilkkionginta}) & \textbf{37.5\%} & 16.7\% \\
50 & Rotary Club activity (\textit{rotarytoiminta}) & 37.5\% & 35.4\% \\
51 & Selkämeri NCO Women (\textit{Selkämeren Alipäällystönaiset ry}) & 37.5\% & 25.0\% \\
52 & Playing chess (\textit{shakinpeluu}) & 37.5\% & 25.0\% \\
53 & Folk dances (\textit{tanhut}) & 37.5\% & 45.8\% \\
54 & Täkänä weaving (\textit{täkänäin kudonta}) & 37.5\% & 18.8\% \\
55 & Viipuri Guild (\textit{Viipurin Kilta ry}) & 37.5\% & 25.0\% \\
56 & Visuvesi Workers' Assoc. (\textit{Visuveden Työväenyhdistys}) & 37.5\% & 50.7\% \\
57 & Horse breeding (\textit{hevoskasvatus}) & 29.2\% & 29.2\% \\
58 & Baltic Sea Summer (\textit{Itämeren Kesä}) & 29.2\% & 36.8\% \\
59 & Agrarian League (\textit{Maalaisliitto}) & 29.2\% & 36.8\% \\
60 & Professional Drivers' Union (\textit{Ammattiautoilijaliitto}) & 25.0\% & 27.1\% \\
61 & Balalaika playing (\textit{balalaikan soittaminen}) & \textbf{25.0\%} & 0.0\% \\
62 & Wilderness Scouts (\textit{Eränkävijät}) & 25.0\% & 50.0\% \\
63 & Heinola Rural Road Board (\textit{Heinolan mlk:n tienhoitokunta}) & 25.0\% & 33.3\% \\
64 & Skiing (\textit{Hiihto}) & 25.0\% & 12.5\% \\
65 & Janakkala Society (\textit{Janakkala-seura}) & 25.0\% & 47.9\% \\
66 & Fabric weaving (\textit{Jcankaankutominen}) & 25.0\% & 75.0\% \\
67 & Cabbage cultivation (\textit{kaalinviljely}) & 25.0\% & 27.1\% \\
68 & Property Inspection Board (\textit{Kiinteistötarkastuslautakunta}) & 25.0\% & 41.7\% \\
69 & Domestic chores (\textit{kodintyöt}) & 25.0\% & 41.7\% \\
70 & Making 3D pictures (\textit{kolmiulotteisten taulujen tekeminen}) & 25.0\% & 12.5\% \\
71 & Typing (\textit{konekirjoitus}) & 25.0\% & 12.5\% \\
72 & Household chores (\textit{kotitaloustyöt}) & 25.0\% & 20.8\% \\
73 & Jogging (\textit{lenkkeily}) & 25.0\% & 6.2\% \\
74 & Salmon farming (\textit{lohen kasvatus}) & 25.0\% & 62.5\% \\
75 & Casual reading (\textit{lueskeleminen}) & \textbf{25.0\%} & 0.0\% \\
76 & Farmers' Assoc. sewing (\textit{Maamiesseuran ompeluseurassa käynti}) & 25.0\% & 41.0\% \\
77 & Rag rug weaving (\textit{matonkutominen}) & 25.0\% & 18.8\% \\
78 & Beekeeping (\textit{mehiläisten hoito}) & 25.0\% & 8.3\% \\
79 & Study circle (\textit{Opintokerho}) & 25.0\% & 27.1\% \\
80 & Lace crocheting (\textit{pitsinvirkkaus}) & 25.0\% & 12.5\% \\
81 & Norden Association (\textit{Pohjola-Norden}) & 25.0\% & 16.7\% \\
82 & Polio Disabled Association (\textit{Polioinvalidit ry}) & 25.0\% & 12.5\% \\
83 & Women of Foremen's Assoc. (\textit{Rakennusmestariyhdistyksen Naisjaosto}) & 25.0\% & 22.9\% \\
84 & Race walking (\textit{ratakävely}) & 25.0\% & 45.8\% \\
85 & Game management (\textit{riistanhoito}) & 25.0\% & 47.9\% \\
86 & Making ryijy rugs (\textit{ryijyjen tekeminen}) & 25.0\% & 22.9\% \\
87 & Rambling/hiking (\textit{samoilu}) & 25.0\% & 31.2\% \\
88 & Knitting socks (\textit{sukkien neulominen}) & \textbf{25.0\%} & 0.0\% \\
89 & Finnish Real Estate Federation (\textit{Suomen Kiinteistöliitto}) & 25.0\% & 41.7\% \\
90 & Finland--Soviet Union Society (\textit{Suomi-Neuvostoliitto-Seura}) & 25.0\% & 54.2\% \\
91 & Studying stars/astronomy (\textit{tutkii tähtiä}) & 25.0\% & 18.8\% \\
92 & Workers' drama club (\textit{Työväenyhdistyksen näytelmäkerho}) & 25.0\% & 18.8\% \\
93 & Vaasa Reserve Officers' Club (\textit{Vaasan Reserviupseerikerho}) & 25.0\% & 27.1\% \\
94 & Sewing clothes (\textit{vaatteiden ompelu}) & 25.0\% & 6.2\% \\
95 & Carving course (\textit{veistokurssi}) & 25.0\% & 43.8\% \\
96 & Viipuri Workers' Institute (\textit{Viipurin Työväenopisto}) & 25.0\% & 37.5\% \\
97 & Violin playing (\textit{viulun soitto}) & 25.0\% & 12.5\% \\
98 & Imatra Region Co-op Bank (\textit{Imatran seudun Osuuspankki}) & 16.7\% & 34.0\% \\
99 & Agricultural Club Association (\textit{Maatalouskerhoyhdistys}) & 16.7\% & 38.2\% \\
100 & Furniture making (\textit{huonekalujen valmistus}) & 12.5\% & 12.5\% \\
101 & Poultry keeping (\textit{kananhoito}) & 12.5\% & 16.7\% \\
102 & Karelia Athletes (\textit{Karjalan Urheilijat}) & 12.5\% & 8.3\% \\
103 & Javelin throw (\textit{keihäänheitto}) & 12.5\% & 0.0\% \\
104 & Home gardening (\textit{kotipuutarhanhoito}) & 12.5\% & 27.1\% \\
105 & Kuusaa Horticultural Assoc. (\textit{Kuusaan Puutarhayhdistys}) & 12.5\% & 40.3\% \\
106 & Ship Officers' Association (\textit{Laivanpäällystöliitto}) & 12.5\% & 33.3\% \\
107 & Lappeenrannan evankelisluterilaisen seurakunta & 12.5\% & 33.3\% \\
108 & Land Acquisition Board (\textit{maanlunastuslautakunta}) & 12.5\% & 39.6\% \\
109 & Seamen's Mission (\textit{merimieslähetys}) & 12.5\% & 49.7\% \\
110 & Metalworkers' Union (\textit{Metallityöväen Liitto}) & 12.5\% & 20.8\% \\
111 & Bankers' Union (\textit{Pankkimiesliitto ry}) & 12.5\% & 33.3\% \\
112 & Paper Workers' Guild (\textit{Paperiammattikunta}) & 12.5\% & 25.0\% \\
113 & Philatelic club (\textit{Postimerkkikerho}) & 12.5\% & 20.8\% \\
114 & Carpentry workshop (\textit{puusepänverstas}) & 12.5\% & 37.5\% \\
115 & Construction Workers' Union (\textit{Rakennustyöväen Liitto}) & 12.5\% & 27.1\% \\
116 & Railway Workers' Choir (\textit{rautatieläisten kuoro}) & 12.5\% & 0.0\% \\
117 & Railway Workers' Union (\textit{Rautatieläisten liitto}) & 12.5\% & 27.1\% \\
118 & Restaurant Staff Union (\textit{Ravintolahenkilökunnan Liitto}) & 12.5\% & 27.1\% \\
119 & Taxation Board (\textit{taksoituslautakunta}) & 12.5\% & 22.9\% \\
120 & Playing tennis (\textit{tenniksen pelaaminen}) & 12.5\% & 14.6\% \\
121 & VR Hyvinkää Men's Choir (\textit{VR:n Hyvinkään mieskuoro}) & 12.5\% & 13.9\% \\
122 & Biathlon (\textit{ampumahiihto}) & 0.0\% & 14.6\% \\
123 & Church Council of Haapavesi (\textit{Haapaveden kirkkoneuvosto}) & 0.0\% & 0.0\% \\
124 & Hämeen Tarmo Women Gymnasts (\textit{Hämeen Tarmon Naisvoimistelijat}) & 0.0\% & 6.2\% \\
125 & Imatra Orienteers (\textit{Imatran Suunnistajat}) & 0.0\% & 12.5\% \\
126 & Bandy (\textit{jääpallo}) & 0.0\% & 0.0\% \\
127 & Primary School Board (\textit{Kansakoululautakunta}) & 0.0\% & 20.1\% \\
128 & Primary school board (\textit{kansakoululautakunta}) & 0.0\% & 18.8\% \\
129 & Kemi Rural Labour Committee (\textit{Kemin mlk:n työvoimatoimikunta}) & 0.0\% & 25.0\% \\
130 & Basketball and Volleyball (\textit{kori- ja lentopalloilu}) & 0.0\% & 0.0\% \\
131 & Municipal Council (\textit{Kunnanvaltuusto}) & 0.0\% & 18.8\% \\
132 & Assoc. Municipal Officials (\textit{Kuntayhtymäin virkamiesyhdistys ry}) & 0.0\% & 31.2\% \\
133 & Singing Masters (\textit{Laulavat mestarit}) & 0.0\% & 0.0\% \\
134 & Choir (\textit{Laulukuoro}) & 0.0\% & 0.0\% \\
135 & Lokalahti Men's Choir (\textit{Lokalahden Laulumiehet}) & 0.0\% & 0.0\% \\
136 & Forestry (\textit{metsänhoito}) & 0.0\% & 31.2\% \\
137 & Men's choir (\textit{mieslaulajiin}) & 0.0\% & 0.0\% \\
138 & Women Gymnasts (\textit{Naisvoimistelijat}) & 0.0\% & 6.2\% \\
139 & Drama club activity (\textit{näytelmäkerhotoiminta}) & 0.0\% & 12.5\% \\
140 & Stage activities/theatre (\textit{näyttämötoiminta}) & 0.0\% & 12.5\% \\
141 & Ball sports (\textit{palloilu}) & 0.0\% & 12.5\% \\
142 & Pukkila Municipal Council (\textit{Pukkilan kunnanvaltuusto}) & 0.0\% & 12.5\% \\
143 & Radio listening (\textit{radion kuuntelu}) & 0.0\% & 0.0\% \\
144 & Mixed choir (\textit{sekakuoro}) & 0.0\% & 0.0\% \\
145 & Non-fiction literature (\textit{tietopuolinen kirjallisuus}) & 0.0\% & 6.2\% \\
146 & Workers' Athletes (\textit{Työväen Urheilijat}) & 0.0\% & 0.0\% \\
147 & Vakka Transport (\textit{Vakka-Kuljetus}) & 0.0\% & 54.2\% \\
148 & Boat building (\textit{veneiden teko}) & 0.0\% & 18.8\% \\
149 & Studying Russian (\textit{venäjänkielen opiskelu}) & 0.0\% & 22.9\% \\
150 & Viljakkala Sports Club (\textit{Viljakkalan urheiluseura}) & 0.0\% & 0.0\% \\\hline
\caption{Model and human disagreement over the 150 test set entities.}
\end{longtable}





\end{document}